\documentclass[11pt,a4paper]{article}
\usepackage[hyperref]{acl2020}
\usepackage{times}
\usepackage{latexsym}

\usepackage{microtype}
\usepackage{graphicx}
\usepackage[absolute,showboxes,overlay]{textpos}
\TPMargin{3pt}

\aclfinalcopy

\title{ur-iw-hnt at GermEval 2021:\\ An Ensembling Strategy with Multiple BERT   Models}

\author{Hoai Nam Tran \\
  Information Science \\
  University of Regensburg \\
  Regensburg, Germany \\
  \texttt{Hoai-Nam.Tran@student.ur.de} \\\And
  Udo Kruschwitz \\
  Information Science \\
  University of Regensburg \\
  Regensburg, Germany \\
  \texttt{Udo.Kruschwitz@ur.de} \\}

\date{}

\begin{document}
\maketitle
\begin{abstract}
This paper describes our approach (ur-iw-hnt) for the Shared Task of GermEval2021 to identify toxic, engaging, and fact-claiming comments. We submitted three runs using an ensembling strategy by majority (hard) voting with multiple different BERT models of three different types: German-based, Twitter-based, and multilingual models. All ensemble models outperform single models, while BERTweet is the winner of all individual models in every subtask. Twitter-based models perform better than GermanBERT models, and multilingual models perform worse but by a small margin.
\end{abstract}

\begin{textblock*}{17cm}[-0.1,0](0cm,27.4cm)
    \centering
    \small
    This work was published as part of the conference proceedings of the GermEval 2021 Workshop on the Identification of Toxic, Engaging, and Fact-Claiming Comments available online at DOI: \href{https://dx.doi.org/10.48415/2021/fhw5-x128}{10.48415/2021/fhw5-x128}. \\
    Please cite as: Hoai Nam Tran and Udo Kruschwitz. ur-iw-hnt at GermEval 2021: An Ensembling Strategy with Multiple BERT Models. In \textit{Proceedings of the GermEval 2021 Workshop on the Identification of Toxic, Engaging, and Fact-Claiming Comments : 17th Conference on Natural Language Processing KONVENS 2021}, pages 83--87, Online (2021).
\end{textblock*}

\section{Introduction}

Moderation of popular social media networks is a difficult task. Facebook alone has almost 2.8 billion active users on April 2021 \citep{kemp_2021}. Moderating discussions between users simultaneously all day is an impossible task, so moderators need help with this work. Also, fully automated solutions for content moderation are not possible, and human input is still required \citep{CambridgeConsultants2019}. An AI-based helper solution for harmful content detection is needed to make social networking less toxic and more pleasant instead.

The Shared Task of GermEval2021 focuses on highly relevant topics for moderators and community managers to moderate online discussion platforms \citep{Risch2021}. The challenge is not to specialize in one broad NLP task like harmful content detection but to detect other essential categories like which comments are engaging or fact-claiming.

We participated in all three subtasks (toxic, engaging and fact-claiming comment classification) to test our ensemble model to see whether multiple BERT-based models provide robust performance for different tasks without further customization. Moderators would benefit from a working system without having to change models or settings all the time.

This report discusses in detail the three runs we submitted in the GermEval2021 Shared Task \citep{Risch2021}. We start with a brief reflection on related work, only focussing on aspects that are closely aligned with the subtasks. We then explain the dataset and the shared tasks in more detail. Next, we present our experiments, some discussions of the results, and we finally draw some conclusions.

To encourage reproducibility of experimental work, we make all code available via GitHub\footnote{\url{https://github.com/HN-Tran/GermEval2021}}.

\section{Related Work}
Detecting harmful content in  social media platforms is not only a monolingual but a multilingual issue. A multilingual toxic text detection classifier uses a fusion strategy  employing  mBERT and XLM-RoBERTa on imbalanced sample distributions \citep{Song2021}. Deep learning ensembles also show their effectiveness in hate speech detection \citep{Zimmerman2019}. A taxonomy of engaging comments contains different possible classifications \citep{Risch2020}. With the increasing spread of misinformation, more collaborations with IT companies specialized in fact-checking and more intelligence and monitoring tools are available to help to identify harmful content \citep{FullFact}. An attempt to fully automate fact-checking is the tool called ClaimBuster \citep{Hassan2015}. Another tool named CrowdTangle monitors social media platforms and alerts the user if specific keywords are triggered so manual fact-claim checking can be done \citep{FullFact}. In addition, an annotation schema for claim detection is also available \citep{Konstantinovskiy2021}. 

\section{Dataset \& Shared Task}
The dataset for the Shared Task of GermEval2021 consists of 3,244 annotated user discussion comments from a Facebook page of the German news broadcast in the timeframe of February to July 2019, labeled by four annotators in three different categories for binary classification: Toxic comments, engaging comments and fact-claiming comments \citep{Risch2021}.
Since the labels are imbalanced, we first applied a stratified split onto the dataset so that 80\% is for training.
We then again apply a stratified split on what is left into two halves, the first part is the development set, and the second part is the holdout set for evaluation which we call the evaluation set here. After training, the ensemble strategy predicts the test dataset, consisting of 944 comments.
%and is later published after the end of the competition. 
Table \ref{dataset-table} shows the imbalance in favor of the negative label. The organizers of GermEval2021 chose the metric Krippendorff's alpha to check each task's intercoder reliability \citep{Risch2021}.

\begin{table}
\centering
\scalebox{0.8}{
    \begin{tabular}{llccc}
        \hline \textbf{Dataset} & \textbf{Label} & \textbf{Subtask 1} & \textbf{Subtask 2} & \textbf{Subtask 3} \\ \hline
        Training & 0 & 2122 & 2379 & 2141\\
        & 1 & 1122 & 865 & 1103 \\ \hline
        Test & 0 & 594 & 691 & 630 \\
        & 1 & 350 & 253 & 314 \\
        \hline
    \end{tabular}}
\caption{\label{dataset-table} Provided training and test dataset}
\end{table}

\subsection{Toxic Comment Classification}
Toxic comments include many harmful and dangerous offenses like "hate speech, insults, threats, vulgar advertisements and misconceptions about political and religious tendencies" \citep{Song2021}. Such behavior only leads to users leaving the discussion or manual bans by the moderator, which can be overwhelming depending on the number of active toxic users \citep{Risch2020}.
For this subtask, the annotator agreement in the usage of insults, vulgar and sarcastic language is $0.73 < \alpha < 0.89$, and in the discrimination, discredition, accusations of lying or threats of violence, the agreement is at $0.83 < \alpha < 0.90$ \citep{Risch2021}. 

\subsection{Engaging Comment Classification}
Engaging comments are, in general, attractive for users to participate in online discussions and get more interactions with other online users in the form of replies and upvotes. A taxonomy of engaging comments has been proposed to identify these comments for detection and classification, so moderators and community managers can reward these comments or posts \citep{Risch2020}.
This task has three different categories \citep{Risch2021}: 
\begin{itemize}
    \item Juristification, solution proposals, sharing of personal experiences ($0.71 < \alpha < 0.89$)
    \item Empathy with regard to other users' standpoints ($0.79 < \alpha < 0.91$)
    \item Polite interaction, mutual respect, mediation ($0.85 < \alpha < 1$)
\end{itemize}

\subsection{Fact-Claiming Comment Classification}
Detecting factual claims is part of the fact-checking process \citep{Konstantinovskiy2021, Babakar2016,DBLP:journals/corr/abs-2103-07769}. The challenge here is to identify claims that have not been fact-checked before and go beyond one sentence that fits into this subtask \citep{Babakar2016}.
Annotator's agreement in fact assertion and evidence provision is at $0.73 < \alpha < 0.84$ \citep{Risch2021}.

\section{Experiments}

\subsection{System Architecture}

\begin{figure*}[!ht]
    \centering
    \includegraphics[width=\textwidth]{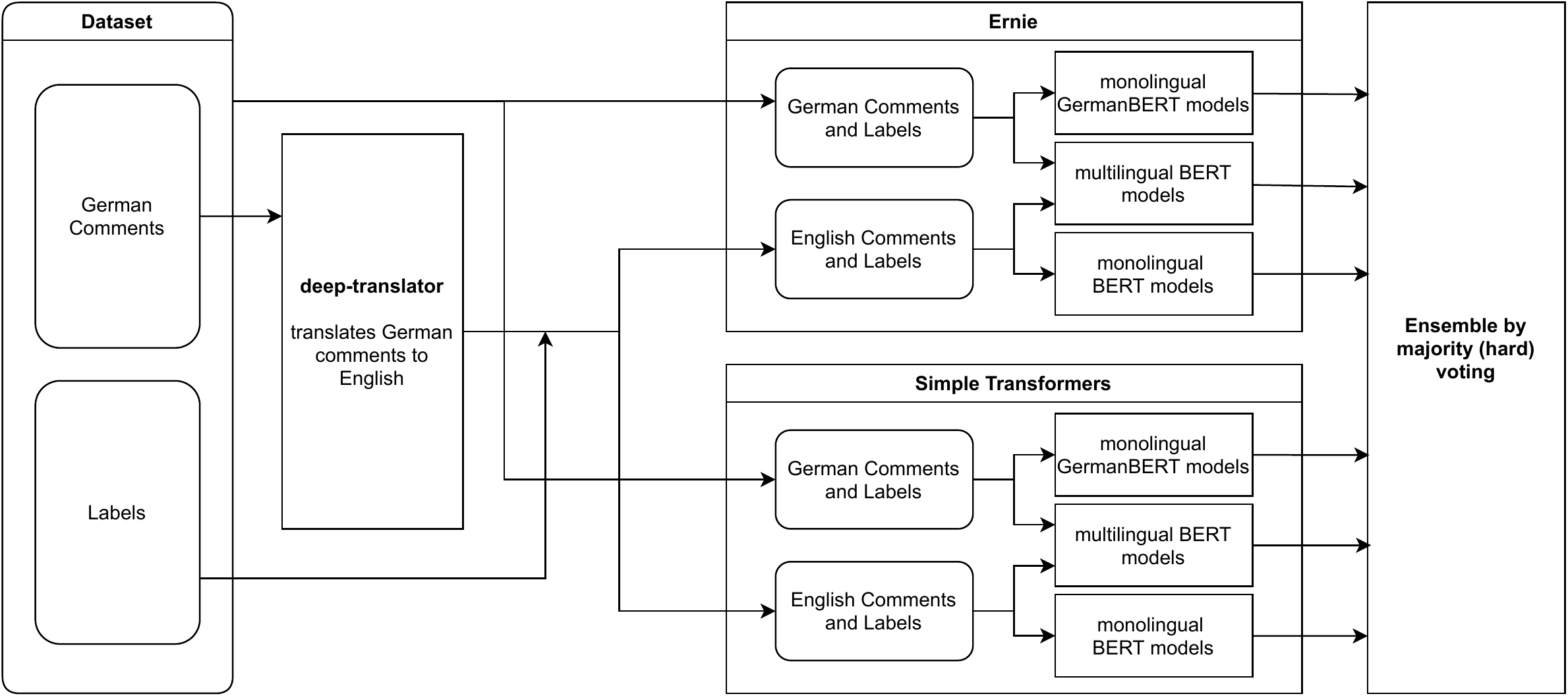}
    \caption{System architecture}
    \label{fig:system}
\end{figure*}

For our system architecture (see Figure \ref{fig:system}), we use three Python libraries/tools. Deep-Translator\footnote{\url{https://github.com/nidhaloff/deep-translator}} translates all the German comments into English by choosing an external service, in our case, the free public Google Translate service. We use two different libraries for classification: Ernie\footnote{\url{https://github.com/labteral/ernie}} and Simple Transformers\footnote{\url{https://simpletransformers.ai/}}. Both work on different versions of HuggingFace's Transformers \citep{Wolf2019} and thus differently: Ernie is a beginner-friendly library last updated in 2020, based on Keras / TensorFlow 2, and uses the optimizer Adam \citep{Kingma2015}. Simple Transformers is based on PyTorch and has more extensive options for hyperparameter tuning and training customizations with AdamW \citep{Loshchilov2019} as the default optimizer. The default 
%or for BERT recommended 
hyperparameter values for our experiments, as recommended for BERT, are in Table \ref{hyperparam-table}.
The only pre-processing step is the tokenization by each BERT model using these libraries. Because of time constraints, cross-validation has not been conducted. After training and evaluating the development and holdout set, the chosen models' predictions go to the ensemble strategy, which finally predicts the test dataset by majority (hard) voting.

\begin{table}
\centering
\scalebox{0.8}{
    \begin{tabular}{lcc}
        \hline \textbf{Hyperparameter} & \textbf{Ernie} & \textbf{Simple Transformers}\\ \hline
        \# epochs & 3 & 3\\
        max sequence length & 128 & 128\\
        learning rate & 2e-5 & 4e-5 \\
        optimizer & Adam & AdamW \\
        \hline
    \end{tabular}}
\caption{\label{hyperparam-table} Hyperparameter values}
\end{table}

\subsection{BERT and its variants}
BERT, which stands for Bidirectional Encoder Representations from Transformers, is a language model developed by Google and is known for its state-of-the-art (SOTA) performance in several NLP tasks \citep{Devlin2019}. The Shared Task consists of German Facebook comments, so we see it fit to choose German-based and English-translation-based models. Because Facebook comments have some similarity with Twitter comments, we also decide on Twitter-based models.

There are several versions of BERT with different pre-training or fine-tuning:
\begin{itemize}
    \item German-based BERT models
    \begin{itemize}
        \item DBMDZ GermanBERT\footnote{\url{https://huggingface.co/dbmdz/bert-base-german-cased}}
        \item Deepset.AI GermanBERT \citep{chan-etal-2020-germans}
    \end{itemize}
    \item Multilingual BERT models
    \begin{itemize}
        \item mBERT Cased \citep{Devlin2019}
        \item XLM-RoBERTa \citep{Conneau2019}
    \end{itemize}
    \item Twitter-based BERT models
    \begin{itemize}
        \item BERTweet \citep{Nguyen2020}
        \item XLM-T \citep{Barbieri2021}
    \end{itemize}
\end{itemize}

Table \ref{bert-table} shows the result of each BERT model on the evaluation/holdout set and on the test dataset with its labels for subtask 1 (which was provided after the submissions had been received).

\begin{table*}
\centering
    \begin{tabular}{llccc}
        \hline \textbf{Classifier} & \textbf{Language} & \textbf{macro F1$_{eval}$} & \textbf{macro F1$_{test}$} \\ \hline
        1) BERT$_{base}$ Uncased \citep{Devlin2019} & English & .6493 & .6329 \\
        2) mBERT$_{base}$ Cased \citep{Devlin2019}& English & .6247 & .6194 \\
        3) mBERT$_{base}$ Cased \citep{Devlin2019}& German & .6286 & .6086 \\
        4) DBMDZ GermanBERT\footnotemark[5] & German & .6472 & .6591 \\
        5) Deepset.AI GermanBERT \citep{chan-etal-2020-germans} & German & .6481 & .6608 \\
        6) BERTweet \citep{Nguyen2020} & English & \textbf{.6798} & \textbf{.6832} \\
        7) XLM-T \citep{Barbieri2021} & English & .6553 & .6681 \\
        8) XLM-T \citep{Barbieri2021} & German & .6342 & .6502 \\
        9) XLM-R$_{base}$ \citep{Conneau2019} & English & .6421 & .6482 \\
        10) XLM-R$_{base}$ \citep{Conneau2019} & German & .3959 & .3862 \\
        \hline
    \end{tabular}
\caption{\label{bert-table} BERT classifier result for subtask 1}
\end{table*}

\subsection{Ensembling Strategy}
The Ensemble Technique is a combination of classifiers' predictions for further classification \citep{Opitz1999}. There are two popular types of ensembling: Bagging \citep{Breiman1996} and Boosting \citep{Schapire}. Ensembles have been shown to be highly effective for a variety of NLP tasks, e.g., in the current top 10 of SQuAD 2.0\footnote{\url{https://rajpurkar.github.io/SQuAD-explorer/}}, all models are ensembles. We went for simple majority ensembling using hard voting, which classifies with the largest sum of predictions from all models.

We decided to use the three runs for the Shared Task to test different combinations of BERT models for a robust and consistent result in the test dataset. That is why we chose five models for the first run, seven models for the second run, and for the third run, nine models ensembled together. The first ensemble consists of two GermanBERT models, the English BERT$_{base}$ model, one Twitter-based model (BERTweet), and one multilingual model, so we have diversity for classification. For the second ensemble, one multilingual model and one Twitter-based model are added. The third ensemble has every classifier except the last one.

The results for each subtask are in Tables \ref{ensemble-table-1}, \ref{ensemble-table-2}, and \ref{ensemble-table-3}, with precision, recall, and macro-averaged F1 score as the scoring metrics. The numbers in the column "Ensemble" refer to the classifier numbers from Table \ref{bert-table}.

\begin{table}[h!]
\centering
\scalebox{0.8}{
    \begin{tabular}{llccc}
        \hline \textbf{Run} & \textbf{Ensemble} & \textbf{P$_{test}$} & \textbf{R$_{test}$} & \textbf{macro F1$_{test}$} \\ \hline
        1 & 1,3,4,5,6 &  .7047 & .6588 & .6810 \\
        2 & 1,2,3,4,5,6,8 &  \textbf{.7183} & \textbf{.6635} & \textbf{.6898} \\
        3 & 1,2,3,4,5,6,7,8,9 & .7168 & .6529 & .6833 \\
        \hline
    \end{tabular}}
\caption{\label{ensemble-table-1} Ensemble result for subtask 1}
\end{table}

\begin{table}[h!]
\centering
\scalebox{0.8}{
    \begin{tabular}{llccc}
        \hline \textbf{Run} & \textbf{Ensemble} & \textbf{P$_{test}$} & \textbf{R$_{test}$} & \textbf{macro F1$_{test}$} \\ \hline
        1 & 1,3,4,5,6 & \textbf{.7228} & \textbf{.6653} & \textbf{.6929} \\
        2 & 1,2,3,4,5,6,8 & .7124 & .6642 & .6875 \\
        3 & 1,2,3,4,5,6,7,8,9 & .7003 & .6542 & .6764\\
        \hline
    \end{tabular}}
\caption{\label{ensemble-table-2} Ensemble result for subtask 2}
\end{table}

\begin{table}[h!]
\centering
\scalebox{0.8}{
    \begin{tabular}{llccc}
        \hline \textbf{Run} & \textbf{Ensemble} & \textbf{P$_{test}$} & \textbf{R$_{test}$} & \textbf{macro F1$_{test}$} \\ \hline
        1 & 1,3,4,5,6 &  \textbf{.7791} & .7310 & .7543 \\
        2 & 1,2,3,4,5,6,8 & .7756 & \textbf{.7454} & \textbf{.7602} \\
        3 & 1,2,3,4,5,6,7,8,9 & .7725 & .7438 & .7579\\
        \hline
    \end{tabular}}
\caption{\label{ensemble-table-3} Ensemble result for subtask 3}
\end{table}

\section{Discussion}
Our experiments demonstrate that BERTweet was showing better performance than every other model in every subtask, which is a surprise. We expected the monolingual GermanBERT models to perform best because of the cultural context in the integrated German language. Multilingual BERT models perform worst but by a close margin. Because of an overfitting issue, the tenth BERT classifier XLM-R performed faultily, only recognizing negative labels and thus the low macro-averaged F1 scores. The margin of each ensemble performance in subtasks 1 and 3 is around 1\%, and for subtask 2 only around 2\%. We conclude that the ensembling strategy shows robustness and consistency for the choice of good classifiers in a big enough amount for each task, and it could be a legitimate approach for the overfitting problem. Because of time constraints, no cross-validation was conducted, and since the holdout set was chosen not to be released for training, there is still improvement in the training quality of the BERT models so that more experiments are needed. Each part of a system like the GPU influences the training accuracy, so an identical replication is difficult to achieve, leading to different results. That is why reproducibility is not guaranteed, even if a manual seed is set\footnote{\url{https://pytorch.org/docs/stable/notes/randomness.html}}. Also, the amount and the imbalance of the dataset can lead to overfitting and lower scoring.

\section{Conclusion and Future Work}
We presented an ensemble strategy using ten  BERT classifiers, including the use of machine translation, demonstrating robustness across tasks. While ensembles perform best overall, Twitter-based models (using standard BERT hyperparameter values) with translation to English perform best in a single model setting. This observation might change if cross-validation, early stopping, hyperparameter tuning, and other optimization techniques for each model are available for future work.
%We presented an ensemble strategy using ten different BERT classifiers, including machine translation for English models. Our experiment shows that the ensembling technique is robust and the best-performing choice. With default or recommended BERT hyperparameter values, Twitter-based models with translation to English perform best in a single model setting, followed by GermanBERT models, and close are multilingual BERT models. This observation might change if cross-validation, early stopping, hyperparameter tuning, and other optimization techniques for each model are available for future work.

\section*{Acknowledgements}

This work was supported by the project \emph{COURAGE: A Social Media Companion Safeguarding and Educating Students} funded by the Volkswagen Foundation, grant number 95564.

\bibliography{acl2020, anthology}

\begin{thebibliography}{22}
\expandafter\ifx\csname natexlab\endcsname\relax\def\natexlab#1{#1}\fi

\bibitem[{Arnold(2020)}]{FullFact}
Phoebe Arnold. 2020.
\newblock \href {https://fullfact.org/media/uploads/coof-2020.pdf} {The
  challenges of online fact checking}.
\newblock Technical report, Full Fact, London, UK.

\bibitem[{Babakar and Moy(2016)}]{Babakar2016}
Mevan Babakar and Will Moy. 2016.
\newblock \href
  {https://fullfact.org/media/uploads/full_fact-the_state_of_automated_factchecking_aug_2016.pdf}
  {{The State of Automated Factchecking}}.
\newblock Technical report, Full Fact, London, UK.

\bibitem[{Barbieri et~al.(2021)Barbieri, Anke, and
  Camacho{-}Collados}]{Barbieri2021}
Francesco Barbieri, Luis~Espinosa Anke, and Jos{\'{e}} Camacho{-}Collados.
  2021.
\newblock \href {http://arxiv.org/abs/2104.12250} {{XLM-T:} {A} multilingual
  language model toolkit for twitter}.
\newblock \emph{CoRR}, abs/2104.12250.

\bibitem[{Breiman(1996)}]{Breiman1996}
Leo Breiman. 1996.
\newblock \href {https://doi.org/10.1007/BF00058655} {Bagging predictors}.
\newblock \emph{Mach. Learn.}, 24(2):123--140.

\bibitem[{{Cambridge Consultants}(2019)}]{CambridgeConsultants2019}
{Cambridge Consultants}. 2019.
\newblock \href
  {https://www.cambridgeconsultants.com/sites/default/files/uploaded-pdfs/Use
  of AI in online content moderation.pdf} {{Use of AI Online in online content
  moderation}}.

\bibitem[{Chan et~al.(2020)Chan, Schweter, and
  M{\"o}ller}]{chan-etal-2020-germans}
Branden Chan, Stefan Schweter, and Timo M{\"o}ller. 2020.
\newblock \href {https://doi.org/10.18653/v1/2020.coling-main.598}
  {{G}erman{'}s next language model}.
\newblock In \emph{Proceedings of the 28th International Conference on
  Computational Linguistics}, pages 6788--6796, Barcelona, Spain (Online).
  International Committee on Computational Linguistics.

\bibitem[{Conneau et~al.(2019)Conneau, Khandelwal, Goyal, Chaudhary, Wenzek,
  Guzm{\'{a}}n, Grave, Ott, Zettlemoyer, and Stoyanov}]{Conneau2019}
Alexis Conneau, Kartikay Khandelwal, Naman Goyal, Vishrav Chaudhary, Guillaume
  Wenzek, Francisco Guzm{\'{a}}n, Edouard Grave, Myle Ott, Luke Zettlemoyer,
  and Veselin Stoyanov. 2019.
\newblock \href {https://doi.org/10.18653/v1/2020.acl-main.747} {{Unsupervised
  cross-lingual representation learning at scale}}.
\newblock \emph{arXiv}.

\bibitem[{Devlin et~al.(2019)Devlin, Chang, Lee, and Toutanova}]{Devlin2019}
J.~Devlin, Ming-Wei Chang, Kenton Lee, and Kristina Toutanova. 2019.
\newblock Bert: Pre-training of deep bidirectional transformers for language
  understanding.
\newblock In \emph{NAACL-HLT}.

\bibitem[{Freund and Schapire(1999)}]{Schapire}
Yoav Freund and Robert Schapire. 1999.
\newblock A short introduction to boosting.
\newblock \emph{Journal-Japanese Society For Artificial Intelligence},
  14(771-780):1612.

\bibitem[{Hassan et~al.(2015)Hassan, Adair, Hamilton, Li, Tremayne, Yang, and
  Yu}]{Hassan2015}
Naeemul Hassan, Bill Adair, J.~Hamilton, C.~Li, M.~Tremayne, Jun Yang, and Cong
  Yu. 2015.
\newblock {The Quest to Automate Fact-Checking}.
\newblock In \emph{Proceedings of the 2015 computation+ journalism symposium}.

\bibitem[{Kemp(2021)}]{kemp_2021}
Simon Kemp. 2021.
\newblock \href
  {https://datareportal.com/reports/digital-2021-april-global-statshot}
  {Digital 2021 april statshot report - datareportal – global digital
  insights}.

\bibitem[{Kingma and Ba(2015)}]{Kingma2015}
Diederik~P. Kingma and Jimmy~Lei Ba. 2015.
\newblock \href {http://arxiv.org/abs/1412.6980} {{Adam: A method for
  stochastic optimization}}.
\newblock \emph{3rd International Conference on Learning Representations, ICLR
  2015 - Conference Track Proceedings}, pages 1--15.

\bibitem[{Konstantinovskiy et~al.(2021)Konstantinovskiy, Price, Babakar, and
  Zubiaga}]{Konstantinovskiy2021}
Lev Konstantinovskiy, Oliver Price, Mevan Babakar, and Arkaitz Zubiaga. 2021.
\newblock \href {https://doi.org/10.1145/3412869} {Toward automated
  factchecking: Developing an annotation schema and benchmark for consistent
  automated claim detection}.
\newblock \emph{Digital Threats: Research and Practice}, 2(2).

\bibitem[{Loshchilov and Hutter(2019)}]{Loshchilov2019}
Ilya Loshchilov and Frank Hutter. 2019.
\newblock \href {http://arxiv.org/abs/1711.05101} {{Decoupled weight decay
  regularization}}.
\newblock \emph{7th International Conference on Learning Representations, ICLR
  2019}.

\bibitem[{Nakov et~al.(2021)Nakov, Corney, Hasanain, Alam, Elsayed,
  Barr{\'{o}}n{-}Cede{\~{n}}o, Papotti, Shaar, and
  Martino}]{DBLP:journals/corr/abs-2103-07769}
Preslav Nakov, David P.~A. Corney, Maram Hasanain, Firoj Alam, Tamer Elsayed,
  Alberto Barr{\'{o}}n{-}Cede{\~{n}}o, Paolo Papotti, Shaden Shaar, and
  Giovanni Da~San Martino. 2021.
\newblock \href {http://arxiv.org/abs/2103.07769} {Automated fact-checking for
  assisting human fact-checkers}.
\newblock \emph{CoRR}, abs/2103.07769.

\bibitem[{Nguyen et~al.(2020)Nguyen, Vu, and Nguyen}]{Nguyen2020}
Dat~Quoc Nguyen, Thanh Vu, and Anh~Tuan Nguyen. 2020.
\newblock \href {https://doi.org/10.18653/v1/2020.emnlp-demos.2} {{BERTweet:
  {A} pre-trained language model for English Tweets}}.
\newblock In \emph{Proceedings of the 2020 Conference on Empirical Methods in
  Natural Language Processing: System Demonstrations, {EMNLP} 2020 - Demos,
  Online, November 16-20, 2020}, pages 9--14. Association for Computational
  Linguistics.

\bibitem[{Opitz and Maclin(1999)}]{Opitz1999}
David Opitz and Richard Maclin. 1999.
\newblock \href {https://doi.org/10.1613/jair.614} {{Popular Ensemble Methods:
  An Empirical Study}}.
\newblock \emph{Journal of Artificial Intelligence Research}, 11:169--198.

\bibitem[{Risch and Krestel(2020)}]{Risch2020}
Julian Risch and Ralf Krestel. 2020.
\newblock \href {http://arxiv.org/abs/2003.11949} {{Top comment or flop
  comment? Predicting and explaining user engagement in online news
  discussions}}.
\newblock \emph{Proceedings of the 14th International AAAI Conference on Web
  and Social Media, ICWSM 2020}, pages 579--589.

\bibitem[{Risch et~al.(2021)Risch, Stoll, Wilms, and Wiegand}]{Risch2021}
Julian Risch, Anke Stoll, Lena Wilms, and Michael Wiegand. 2021.
\newblock Overview of the {G}erm{E}val 2021 shared task on the identification
  of toxic, engaging, and fact-claiming comments.
\newblock In \emph{Proceedings of the GermEval 2021 Shared Task on the
  Identification of Toxic, Engaging, and Fact-Claiming Comments co-located with
  KONVENS}, pages 1--12.

\bibitem[{Song et~al.(2021)Song, Huang, and Xiao}]{Song2021}
Guizhe Song, Degen Huang, and Zhifeng Xiao. 2021.
\newblock \href {https://doi.org/10.3390/info12050205} {{A study of
  multilingual toxic text detection approaches under imbalanced sample
  distribution}}.
\newblock \emph{Information (Switzerland)}, 12(5):1--16.

\bibitem[{Wolf et~al.(2020)Wolf, Debut, Sanh, Chaumond, Delangue, Moi, Cistac,
  Rault, Louf, Funtowicz, Davison, Shleifer, von Platen, Ma, Jernite, Plu, Xu,
  Le~Scao, Gugger, Drame, Lhoest, and Rush}]{Wolf2019}
Thomas Wolf, Lysandre Debut, Victor Sanh, Julien Chaumond, Clement Delangue,
  Anthony Moi, Pierric Cistac, Tim Rault, Remi Louf, Morgan Funtowicz, Joe
  Davison, Sam Shleifer, Patrick von Platen, Clara Ma, Yacine Jernite, Julien
  Plu, Canwen Xu, Teven Le~Scao, Sylvain Gugger, Mariama Drame, Quentin Lhoest,
  and Alexander Rush. 2020.
\newblock \href {https://doi.org/10.18653/v1/2020.emnlp-demos.6} {Transformers:
  State-of-the-art natural language processing}.
\newblock In \emph{Proceedings of the 2020 Conference on Empirical Methods in
  Natural Language Processing: System Demonstrations}, pages 38--45, Online.
  Association for Computational Linguistics.

\bibitem[{Zimmerman et~al.(2019)Zimmerman, Fox, and Kruschwitz}]{Zimmerman2019}
Steven Zimmerman, Chris Fox, and Udo Kruschwitz. 2019.
\newblock {Improving hate speech detection with deep learning ensembles}.
\newblock \emph{LREC 2018 - 11th International Conference on Language Resources
  and Evaluation}, pages 2546--2553.

\end{thebibliography}
\bibliographystyle{acl_natbib}
\end{document}